\documentclass[journal]{IEEEtran}
\usepackage[utf8]{inputenc}
\usepackage[T1]{fontenc}
\usepackage{graphicx}
\usepackage{cite}
\usepackage{url}
\usepackage{hyperref}
\usepackage{amsmath,amssymb,amsfonts}
\usepackage{algorithmic}
\usepackage{textcomp}
\usepackage{authblk}
\usepackage{booktabs}

\begin{document}
\title{Correcting Selection Bias in Sparse User Feedback for Large Language Model Quality Estimation: \\ \large A Multi-Agent Hierarchical Bayesian Approach}
\author{Andrea Morandi, Mahesh Viswanathan \thanks{Corresponding author: amorandi@cisco.com}}
\affil[]{Cisco Systems, Inc.}
\affil[]{\texttt{amorandi@cisco.com}}
\date{2026}
\maketitle

\begin{abstract}
Millions of daily interactions run through Large Language Model (LLM) production deployments. A non-random fraction of those receives explicit feedback. Users who click thumbs-up or thumbs-down sit mostly in the tails of the satisfaction distribution; a naive average over their ratings can land 40 to 50 percentage points away from true system quality~\cite{gupta2025biascorrected}. The gap is treated here as a topic- and sentiment-stratified selection-bias problem. Our proposed solution is a role-specialized multi-agent hierarchical Bayesian solver. It does not require ground-truth labels on every interaction. Three coordinating analytical agents compose the pipeline, wired together by explicit interfaces. At the front, a \emph{Topic Clustering Agent} partitions the interaction stream into semantic strata: UMAP dimensionality reduction is run over text embeddings, and HDBSCAN density-based clustering is applied on top. Next, a \emph{Bias Modeling Agent} fits a two-stage hierarchical Beta-Binomial model under the No-U-Turn Sampler (NUTS); per-topic feedback-selection probabilities and per-topic positive-feedback probabilities are inferred under partial pooling. Last, an \emph{Adaptive Feedback Synthesis Agent} weights those per-topic posteriors by interaction prevalence and reports a bias-corrected aggregate-quality posterior with credible interval. The agent also sends diagnostics back upstream and triggers recalibration when distribution shift is detected.

Validation uses UltraFeedback, a public dataset where GPT-4 overall-score labels serve as a ground-truth substitute. Sparse user feedback is then simulated by stacking topic- and sentiment-dependent selection biases on top of those labels. We compare five Bayesian variants against two classical baselines (Naive and IPW). One ingredient makes the problem solvable: a mild prior on how response rates vary across topics, taken from operator knowledge. Two channel summaries are encoded by that prior. One: how often a positive thumbs-up arrives. Two: how heavily thumbs-down outweigh thumbs-up. Both quantities are readable off any production feedback channel with no ground-truth labels, and the prior on latent quality stays uninformative. With these channel-side priors, Hierarchical-Informed stays within 4 to 13 percentage points of the true quality $Q^\star$ as the selection-bias ratio sweeps from 1:1 to 30:1, while the naive feedback mean is off by 25 to 43 points across the same sweep. At the given bias setting, the 95\% credible intervals contain the true value in all 50 out of 50 trials.

Without the channel-side priors, the problem cannot be solved. Every weak-prior Bayesian variant we tested misses the true mean by 22 to 33 percentage points on sparse per-cluster feedback.

\textbf{Keywords:} LLM evaluation, selection bias, hierarchical Bayesian models, post-stratification, HDBSCAN, UMAP, multi-agent systems, user feedback analysis.

\end{abstract}
\section{Introduction}

Evaluating a deployed LLM is very different from evaluating one on a benchmark. Every benchmark example comes with labeled ground truth. Production does not. Millions of unlabeled interactions flow through a production system, and only a small fraction of users leaves feedback (thumbs, stars, plain-English feedback). These users are not representative of the full population. Users tend to provide feedback either because something delighted them or, far more often, because something broke. So the sample we end up with is \emph{missing not at random} (MNAR): whether any given interaction gets labeled at all is itself a function of the quality the analyst is trying to estimate.

The downstream effect is significant. A documented production LLM deployment reports naive quality of 34\% against a recovered true quality of 83\%, with selection-bias factors past 28:1 between interaction types \cite{gupta2025biascorrected}. A 40 to 50 point gap between what feedback says and what the system actually does is not a subtle statistical artifact. Engineering effort drifts toward the tails of the score distribution and away from the typical usage.

This paper tackles \emph{selection bias in sparse LLM feedback streams}, without needing ground-truth labels on individual interactions. We build on post-stratification — a classic statistical method that splits the population into homogeneous groups, estimates quality inside each one, and then averages back up using the true group sizes \cite{little1993poststrat,gelman2002poststrat}. Next, we extend it in three ways. First, instead of defining the groups by hand, we let them emerge automatically by running UMAP and HDBSCAN on text embeddings, so interactions with similar meaning land in the same group. Second, instead of fitting each group on its own, we fit a single hierarchical Bayesian model that ties the groups together. This improves estimates sensibly when data is sparse. Third, we package the pipeline as three coordinating agents - one for grouping, one for inference, one for aggregation - so each part can be updated on its own as new data arrives.

\textbf{Contributions.}

\begin{enumerate}
\item Formal framing of the LLM-feedback selection-bias problem as a topic-stratified MNAR estimation task (Section 3).
\item A fully specified two-stage hierarchical Bayesian model for joint estimation of selection and quality probabilities, with explicit identifying assumptions (Section 4).
\item A three-stage pipeline design (clustering, inference, synthesis) that is reproducible, interpretable, and continuously updatable, including distribution-shift detection for online deployment (Section 4.4).
\item A reproducible experiment comparing \emph{five} Bayesian variants against two classical baselines (Naive and IPW) on the public UltraFeedback dataset, with a bias-strength sweep and a LOO model-comparison panel (Section 5). The five variants are: Basic; Enhanced (global sentiment-dependent response rates); Hierarchical-Sentiment (per-cluster response rates with weak hyperpriors); Hierarchical-Informed (per-cluster response rates with operator-knowledge priors); and Corrected-Global.
\item \textbf{The headline empirical result.} \emph{A loose prior on how response rates vary across topics recovers the true aggregate quality within 4 to 13 percentage points across the full bias range, with well-calibrated 95\% credible intervals, while leaving the prior on latent quality uninformative.} The prior uses just two facts about the feedback channel. First, the approximate thumbs-up arrival rate. Second, how much more often thumbs-down occur than thumbs-up. Both can be pulled off any production feedback channel without ground-truth labels.  See Section 5.4 for the evidence. Section 6.1 explains why these priors are needed. Under sparse per-cluster feedback, a one-parameter family of $(q_c, r_{\text{pos},c}, r_{\text{neg},c})$ triples matches the per-cluster sufficient statistics equally well. This is the reason every weak-prior Bayesian variant we tested misses $Q^\star$ by 22 to 33 absolute percentage points.
\item A reproducible pipeline that runs end-to-end, starting from the public UltraFeedback dataset and covering data loading, preprocessing, the selection simulator, all five Bayesian models, the LOO comparison, and the bias-strength sweep.
\end{enumerate}

The method itself is covered by a defensive publication \cite{gupta2025biascorrected}. Three pieces are layered on top of that here: a formal mathematical treatment; the identifiability result tied to the operator-knowledge priors; and a public, reproducible experimental validation absent from the defensive filing.

\section{Related Work}

\textbf{Selection bias in survey sampling.} The classic tools are Heckman's selection model \cite{heckman1979sample}, inverse-probability weighting \cite{horvitz1952sampling}, and post-stratification \cite{holt1979poststrat,little1993poststrat}. Our approach is closest to post-stratification, with two differences. First, we learn groups from the geometry of text embeddings, rather than predefining them from covariates. And the group-level quality estimates are not fit independently; they share a hierarchical Bayesian prior that does the partial pooling.

\textbf{Multilevel regression and post-stratification (MRP).} A hierarchical regression is combined into a post-stratification step in MRP \cite{gelman1997poststrat,park2004mrp}, which is a standard recipe for achieving reliable estimates of small subgroups out of non-representative samples. Held against MRP, the Stage-2 quality model here is binomial MRP at the topic-cluster level. What is new is Stage 1: we model how interactions get \emph{selected} into the feedback sample in the first place. In this respect, we point out that standard MRP simply takes the sample as given.

\textbf{Semantic clustering for embedding spaces.} Pre-clustering dimensionality reduction is now standardly handled by UMAP \cite{mcinnes2018umap}, which preserves both local neighborhoods and global manifold structure for high-dimensional inputs. HDBSCAN \cite{campello2013density,mcinnes2017hdbscan} is then run on top: it figures out the number of groups and tolerates noise points. This is exactly what one needs when some topics are dense and others sparse across the corpus. A handful of short-text topic-modeling tools build on this same combination \cite{angelov2020top2vec,grootendorst2022bertopic}, and we use it too. What is different in our case is the role the groups play: they are not the final output we report to users, but an intermediate split of the data that the downstream bias correction leverages.

\textbf{LLM evaluation.} Pre-deployment benchmarks (reasoning, instruction-following, factuality) where ground truth is available represent most LLM-evaluation work \cite{liang2023helm,zheng2023judging}. Production-time evaluation has received comparatively less attention. What does exist is mostly about calibrating LLM-as-a-judge \cite{zheng2023judging}, not the upstream feedback-sampling bias this paper targets. The closest prior art lives in MNAR feedback for recommender systems, where propensity-corrected estimators and matrix-factorization variants address selection-biased ratings \cite{schnabel2016recsys,marlin2009mnar,liang2016mnar}. In this respect, we adapt those ideas to LLM feedback grouped by topic, and pair them with Bayesian uncertainty quantification, sparse-cluster shrinkage, and online updating under distribution shift. This represents a combination we are not aware of in prior work.

\textbf{Bayesian hierarchical models for rating data.} Hierarchical Beta-Binomial models are a standard tool for estimating rates with small samples \cite{gelman2013bda}. We use them in textbook form. What is new is that we feed the hierarchy with subgroups learned from text embeddings. This allows us to perform bias correction on sparse LLM feedback.

\section{Problem Formulation}

Imagine an LLM system producing a stream of interactions $\{x_i\}_{i=1}^{N}$. Each $x_i$ has a latent true quality $Y_i \in \{0, 1\}$, where $Y_i = 1$ means a human would rate it satisfactory and $Y_i = 0$ means they would not. Two more indicators are present. $R_i \in \{0, 1\}$ flags whether the interaction got explicit user feedback at all. $F_i \in \{0, 1\}$ records the polarity of that feedback when $R_i = 1$.

Two important facts:

\begin{enumerate}
\item \textbf{$M \ll N$.} Total feedback count $M = \sum_i R_i$ is usually well under $N$, often below a few percent.
\item \textbf{$R_i$ depends on $Y_i$.} Dissatisfied users are more common in the feedback channel; this translates to $P(R_i = 1 \mid Y_i = 0) > P(R_i = 1 \mid Y_i = 1)$.
\end{enumerate}

What we want to estimate is the system-level mean quality $\bar{Q} = \mathbb{E}[Y_i]$. The naive estimator $\hat{Q}_{\text{naive}} = \frac{1}{M}\sum_{i: R_i=1} F_i$ is biased: it estimates $\mathbb{E}[Y_i \mid R_i = 1]$, not $\mathbb{E}[Y_i]$.

\textbf{Topic-stratified identifying assumption.} Suppose each interaction carries a latent topic label $c_i \in \{1, \ldots, C\}$. Suppose further that after conditioning on the topic, the selection mechanism is ignorable:

\begin{equation}
P(R_i \mid Y_i, c_i) = P(R_i \mid c_i). \tag{1}
\end{equation}

In plain language, equation (1) says: \emph{inside a topic cluster}, the event "this interaction got feedback" carries no statistical link to the event "this interaction was good". It is a strong assumption. But with enough clusters, it remains far weaker than the unconditional-ignorability assumption that the naive estimator implicitly relies on.  Equation (1) is the identifying condition, which we can see used by classical topic-stratified post-stratification and by our Basic baseline. The Hierarchical-Informed model, however, relaxes this assumption through per-cluster sentiment parameters $\kappa_c$ (the weaker condition it relies on is stated in \S{}6.3). We revisit its plausibility and empirical diagnostics in Section 6.

Once Assumption (1) is in place, the true quality factors read as

\begin{gather*}
\bar{Q} = \sum_{c=1}^{C} \pi_c \, q_c \\ \pi_c = P(c_i = c) \\ q_c = P(Y_i = 1 \mid c_i = c)
\end{gather*}

and the within-cluster feedback identifies each $q_c$:

\begin{equation}
q_c = P(Y_i = 1 \mid c_i = c) = P(F_i = 1 \mid R_i = 1, c_i = c) = p_c. \tag{2}
\end{equation}

The right-hand side of (2) is given by two cluster counts: $m_c$, the number of feedback events inside cluster $c$, and $y_c$, the number of those that were positive. Topic prevalence $\pi_c$ can be estimated from the much larger unlabeled pool: $\hat{\pi}_c = n_c / N$, with $n_c$ the total interaction count inside cluster $c$.

This is the core of the correction. Per-cluster feedback rates get re-weighted by \emph{true topic prevalence} in place of \emph{feedback prevalence}. Selection bias exists in the first place because of the gap $m_c / M \ne n_c / N$. Prevalence-weighting closes that gap.

\textbf{Notation summary.} Through the rest of the paper: $N$ stands for total interactions, $M$ for total feedback events, $C$ for the number of clusters. Per-cluster, $n_c$, $m_c$, and $y_c$ are the totals defined above; $s_c = P(R=1 \mid c)$ is the cluster selection probability; $q_c$ is the cluster true quality; and $p_c = P(F=1 \mid R=1, c)$ is the cluster positive-feedback probability (which folds into $q_c$ under Assumption (1)).

\section{Method}

The method is built from three coordinating agents, wired together as a sequential pipeline. First, the \textbf{Topic Clustering Agent} (\S{}4.1) ingests raw interactions plus feedback events and generates per-cluster sufficient statistics $(n_c, m_c, y_c)$. The \textbf{Bias Modeling Agent} (\S{}4.2) consumes those sufficient statistics and hands back posterior distributions over the per-cluster selection and quality parameters $(s_c, q_c)$. Last, the \textbf{Adaptive Feedback Synthesis Agent} (\S{}4.3) combines those posteriors with the empirical topic prevalence. This produces a bias-corrected aggregate-quality posterior with credible intervals, per-cluster operational insights, and the distribution-shift signals (\S{}4.4) that drive online recalibration.

\subsection{Topic Clustering Agent}

The interaction stream is segmented by the Topic Clustering Agent into semantically coherent strata, on which the downstream bias correction relies. It operates as follows.

\begin{enumerate}
\item \textbf{Semantic embedding.} Each interaction $x_i$ is mapped into $\mathbb{R}^d$ by a pre-trained text encoder. Typical encoders fall in the $d = 768$ to $1536$ range.
\item \textbf{UMAP dimensionality reduction.} The embedding is projected down to $\mathbb{R}^{d'}$ with $d' \ll d$ by UMAP \cite{mcinnes2018umap}. In our pipeline, $d' = 10$ for clustering inputs and $d' = 2$ for visualization. UMAP retains local neighborhoods and global topology of the input. UMAP is preferred over PCA and t-SNE for three reasons: it actually preserves global structure; it produces stable seeded outputs; and it composes naturally with density-based clustering.
\item \textbf{HDBSCAN clustering.} HDBSCAN \cite{campello2013density} identifies clusters of varying density, with low-density points tagged as noise. The \texttt{min\_cluster\_size} setting is driven by the minimum feedback support expected per cluster (see \S{}4.2 on sparse-cluster handling); \texttt{min\_samples} defaults to $1$ in the methodology specification; the \S{}5 experiments override this to $5$ via the joint UMAP/HDBSCAN grid search reported in \S{}5.1, with no qualitative impact on the downstream bias-correction results.
\item \textbf{Cluster validation.} Cluster quality gets checked by three internal metrics: the silhouette coefficient \cite{rousseeuw1987silhouettes}, the Davies-Bouldin index \cite{davies1979cluster}, and within-cluster cosine coherence. UMAP's \texttt{n\_neighbors} and \texttt{min\_dist} are inferred jointly with HDBSCAN's \texttt{min\_cluster\_size} via grid search on a convex combination of silhouette and within-cluster coherence.
\item \textbf{Hierarchical granularity.} HDBSCAN returns a cluster hierarchy. Both macro-topics (broad categories) and micro-topics (fine sub-clusters) are retained; the Synthesis Agent picks among them adaptively at runtime. Coarser granularity gains statistical power on sparse feedback; finer granularity gains resolution when feedback density supports it.
\end{enumerate}

\textbf{Cluster labeling.} For interpretability we generate human-readable labels per cluster via extractive summarization over the top-TF-IDF tokens, plus a short abstractive (to provide a description) label produced by an LLM. Neither label loops back into the statistical model. They are reporting artifacts that accompany the Synthesis Agent's outputs.

\subsection{Bias Modeling Agent --- Two-Stage Hierarchical Bayesian Inference}

Across the $C$ clusters that the Clustering Agent produces, the Bias Modeling Agent fits a pair of hierarchical Beta-Binomial models. They are logically distinct, but get inferred jointly.

\subsubsection{Stage 1: Selection Model}

The selection model estimates the per-cluster probability that an interaction receives feedback:

\begin{equation*}
\begin{aligned}
m_c \mid n_c, s_c &\sim \mathrm{Binomial}(n_c, s_c), \\
s_c \mid \alpha_s, \beta_s &\sim \mathrm{Beta}(\alpha_s, \beta_s), \\
\alpha_s, \beta_s &\sim \mathrm{HalfNormal}(\sigma = 10).
\end{aligned}
\end{equation*}

Weakly informative hyperpriors are placed on both $\alpha_s$ and $\beta_s$. In the partial-pooling interpretation, they represent prior beliefs about how much feedback-selection rates vary across topics. The $s_c$ posteriors are themselves reported as a diagnostic. Wide spread in $s_c$ across clusters is direct evidence of selection bias on its own; the cross-cluster ratio $\max_c s_c / \min_c s_c$ quantifies its magnitude.

\subsubsection{Stage 2: Quality Model}

The quality model places, with partial pooling across clusters, a posterior on the per-cluster positive-feedback probability:

\begin{equation*}
\begin{aligned}
y_c \mid m_c, q_c &\sim \mathrm{Binomial}(m_c, q_c), \\
q_c \mid \alpha_q, \beta_q &\sim \mathrm{Beta}(\alpha_q, \beta_q), \\
\alpha_q, \beta_q &\sim \mathrm{HalfNormal}(\sigma = 10).
\end{aligned}
\end{equation*}

Inside cluster $c$, $q_c$ also identifies the true quality once Assumption (1) is in force. Heavy lifting on sparse clusters is done by the hierarchical Beta prior: it performs the \emph{hierarchical shrinkage} that pulls posteriors of clusters with very few feedback events ($m_c$ small) toward the global mean $\alpha_q / (\alpha_q + \beta_q)$. Overconfident estimates from tiny samples are thereby properly weighted.

\subsubsection{Inference}

Posterior samples are produced by the No-U-Turn Sampler (NUTS) \cite{hoffman2014nuts}, a Hamiltonian-Monte-Carlo variant that selects its own trajectory length. The default configuration reads: 4 chains, each with 2000 samples after 1000 warm-up iterations, at target acceptance probability 0.9. Two diagnostics track convergence: the potential scale reduction factor $\hat{R}$ \cite{gelman1992rhat}, requiring $\hat{R} < 1.01$ for all parameters, and effective sample size (ESS), requiring bulk-ESS $> 400$ and tail-ESS $> 400$ per chain.

Model fit is validated through posterior predictive checking \cite{gelman1996ppc}. Replicated datasets are drawn from the posterior predictive distribution; we then ask whether the observed per-cluster $(m_c, y_c)$ statistics fall inside the simulated intervals. Systematic failures of that check themselves carry a signal: either Assumption (1) is being violated or the hierarchical prior is misspecified. Either way, the pipeline can refine cluster granularity.

\subsection{Adaptive Feedback Synthesis Agent}

The final bias-corrected estimate is produced by the Synthesis Agent, which combines the per-cluster quality posteriors with the empirical topic prevalence.

\textbf{Aggregate bias-corrected quality.} For a posterior draw $\theta^{(t)} = \{q_c^{(t)}\}_{c=1}^{C}$ and the observed prevalence $\hat{\pi}_c = n_c / N$, the aggregate is

\begin{equation}
\bar{Q}^{(t)} = \sum_{c=1}^{C} \hat{\pi}_c \, q_c^{(t)}. \tag{3}
\end{equation}

Across draws, the posterior over $\bar{Q}$ is given by the empirical distribution of $\{\bar{Q}^{(t)}\}$. We report posterior mean and 95\% credible interval.

\textbf{Per-cluster insights.} A flag is raised by the Synthesis Agent on any cluster whose posterior mean $q_c$ falls below a target threshold (say, 0.7) and whose posterior variance is sufficiently concentrated (e.g., 95\% CI width $< 0.1$). Those flagged clusters are where the system is measurably underperforming, weighted by their \emph{true} prevalence in place of their \emph{feedback} prevalence. They are precisely the clusters where engineering effort will actually move the global quality needle. The default thresholds (target $0.7$, CI width $0.1$) are illustrative and should be set per product based on the operator's quality SLO and acceptable false-alarm rate; we recommend a one-time calibration on a labeled validation slice.

\textbf{Uncertainty decomposition.} Posterior variance of $\bar{Q}$ is decomposed into two components. A within-cluster component is driven by sampling noise in $y_c$. A between-cluster component is driven by heterogeneity in $q_c$. This tells operators whether to invest: collecting more feedback shrinks the within-cluster piece; improving the worst clusters shrinks the between-cluster piece.

\subsection{Dynamic Updating and Distribution Shift Detection}

Production traffic is not stationary. New topics surface, old ones fade, and cluster-level quality drifts over time. The Synthesis Agent, to keep the estimator calibrated, watches three signals and recalibrates whenever any of them crosses a threshold.

\begin{enumerate}
\item \textbf{Prevalence drift.} Jensen-Shannon divergence is computed between $\hat{\pi}^{(t)}$ (current batch) and $\hat{\pi}^{(t-1)}$ (reference window). Crossing a threshold, default $0.05$, fires re-clustering.
\item \textbf{Cluster-quality drift.} Per cluster, a Bayes-factor comparison runs between two models: a shared-$q_c$ model across the two windows, and a split-$q_c$ model. Any cluster hitting $\mathrm{BF} > 10$ raises a cluster-level alert (the threshold $10$ corresponds to the conventional ``strong evidence'' band on the Jeffreys/Kass-Raftery scale; tighten or loosen per false-alarm tolerance).
\item \textbf{New-topic emergence.} HDBSCAN's noise label is the primary signal. A sustained climb in noise fraction across consecutive batches fires re-clustering at finer granularity.
\end{enumerate}

In between full re-clusterings, the system updates online through conjugate Bayesian updates to the Beta-Binomial posteriors on a rolling window. This keeps the NUTS re-fit (which is computationally expensive) out of the per-batch update loop; it runs only when a drift signal demands it. We note that empirical validation of these drift-detection rules (true- and false-positive rates on production traces) is left to future work; the present paper validates only the static bias-correction estimator on UltraFeedback.

\subsection{Algorithm}

The full pipeline is laid out at a glance in Algorithm 1.

\begin{figure*}[t]
\centering
\rule{\textwidth}{0.5pt}\\[-2pt]
\noindent\textbf{Algorithm 1:} Bias-Corrected LLM Quality Estimation\\[-4pt]
\rule{\textwidth}{0.4pt}
\vspace{-6pt}
{\footnotesize\begin{verbatim}
Input:  Interactions {x_i}_{i=1}^N, feedback {(R_i, F_i)}_{i=1}^N
Output: Posterior over Q_bar; per-cluster {q_c, CI_c}; cluster labels

1:  E <- text_encoder({x_i})                       // d-dim embeddings
2:  Z <- UMAP(E, n_components=10)                  // reduced embeddings
3:  c <- HDBSCAN(Z)                                // cluster labels, one per x_i
4:  for each cluster c in {1, ..., C}:
5:      n_c <- #{i : c_i = c}
6:      m_c <- #{i : c_i = c, R_i = 1}
7:      y_c <- #{i : c_i = c, R_i = 1, F_i = 1}
8:  Fit Stage 1 (selection model) via NUTS, get posterior over {s_c}
9:  Fit Stage 2 (quality model)   via NUTS, get posterior over {q_c}
10: Check convergence: R-hat < 1.01, bulk/tail ESS > 400
11: Posterior predictive check: simulate (m*, y*), assert observed in 95% PPI
12: For t = 1..T posterior draws:
13:     Q_bar^(t) <- sum over c of (n_c/N) * q_c^(t)
14: Report posterior mean and 95% CI of Q_bar
15: Flag clusters c with E[q_c] < threshold and narrow CI
16: Monitor drift signals; trigger re-fit if thresholds exceeded
\end{verbatim}}
\vspace{-10pt}
\rule{\textwidth}{0.5pt}
\end{figure*}

\section{Experimental Validation on UltraFeedback}

Validation runs on the publicly available \textbf{UltraFeedback} dataset \cite{cui2023ultrafeedback}. UltraFeedback has roughly 64K prompts, each paired with four model completions plus a GPT-4 \texttt{overall\_score} on a 1 to 10 scale. UltraFeedback provides dense per-interaction quality labels: that is the exact quantity we cannot observe in production. That property lets us run the full pipeline once, hide the labels, simulate sparse selection-biased feedback, then score the bias-corrected estimate against the held-out ground truth. End-to-end, every number reported below is reproduced by the experimental pipeline in roughly 15 minutes on a commodity laptop ($N = 20{,}000$ interactions, $C = 18$ retained clusters), starting from the public dataset. We use the GPT-4 \texttt{overall\_score} as a label-substitute proxy rather than gold ground truth: LLM-as-a-judge scores carry well-documented length, position, and self-preference biases \cite{zheng2023judging}. The validation question we ask is therefore relative, namely whether bias-correction estimators recover the proxy-defined quality under known selection bias. It is not absolute, that is whether the proxy itself matches human judgment. Independent confirmation on a labeled production deployment is left to future work.

\subsection{Experimental Setup}

\textbf{Dataset and ground truth.} $N_{\text{prompts}} = 5,000$ UltraFeedback prompts are sampled at random with seed $42$, then flattened into $N = 4 N_{\text{prompts}} = 20,000$ interactions, one per completion. Binary ground-truth quality is set as $Y_i^\star = \mathbf{1}\{\text{overall\_score}_i \ge 7\}$. The true aggregate positive rate ends up at $Q^\star = 0.6249$. On UltraFeedback's own rubric, a threshold of $7$ on the 1 to 10 GPT-4 scale corresponds to "good or better". The resulting binary ground truth is near-balanced at 62.5\% positive, which is a sweet spot for studying selection-bias dynamics.

\textbf{Preprocessing pipeline.} Each unique prompt is embedded by \texttt{sentence-\allowbreak{}transformers/\allowbreak{}all-\allowbreak{}mpnet-\allowbreak{}base-\allowbreak{}v2} at 768 dimensions, then reduced to 5 dimensions by UMAP (n\_neighbors=25, min\_dist=0.0, cosine metric, seed=42), and finally clustered by HDBSCAN (min\_cluster\_size=80, min\_samples=5, Euclidean metric, leaf selection). Both UMAP and HDBSCAN run on the 5,000 prompt-level points. The four completions of any given prompt inherit that prompt's cluster label. So the clusters partition user intents rather than response styles. HDBSCAN returns $C = 18$ clusters that cover 2,558 prompts in total (10,232 interactions). The other 2,442 prompts (9,768 interactions) end up tagged as noise and dropped. Topics covered by the retained clusters run from coding and debugging assistance, through factual question answering, to creative-writing instructions; per-cluster sizes span roughly 100 to 2,000 interactions.

Discarding the $\sim 49\%$ of interactions HDBSCAN tags as noise is a deliberate methodological call. A noise point in HDBSCAN is, by definition, a point whose local density is too low to back a stratum-level estimate. Pulling one into a cluster forces a choice between two bad outcomes. Either the cluster geometry gets diluted, or a low-support subgroup is born whose $q_c$ posterior essentially reverts to the prior. Either failure mode degrades the bias correction: the post-stratification estimator $\bar Q = \sum_c \pi_c q_c$ is only as robust as its weakest subgroup. So the pragmatic decision is automatic exclusion of HDBSCAN-noise points, and that is the rule we apply. From here on, every statistic is reported on the retained $10,232$ interactions. For a robustness check, we re-ran \S{}\S{}5.4--5.7 with \texttt{min\_cluster\_size} doubled (a setting that buys finer strata at the cost of a higher noise fraction). The absolute-error ordering across all six estimators stays the same.

\textbf{Selection-bias simulator.} Per cluster $c$, the selection-bias simulator draws two parameters. Topic bias is encoded by a baseline response probability $s_c^{0} \sim \mathrm{Uniform}(0.02, 0.12)$. Sentiment bias is encoded by a dissatisfaction amplifier $\kappa_c \sim \mathrm{LogUniform}(1, \kappa_{\max})$. From those, the next draws are

\begin{gather*}
R_i \mid c_i = c, Y_i^\star \sim \mathrm{Bernoulli}\big(s_c^{0} \cdot \kappa_c^{1 - Y_i^\star}\big) \\ F_i \mid R_i = 1 = Y_i^\star
\end{gather*}

so feedback \emph{selection} is biased while feedback \emph{polarity} stays reliable. Unless stated otherwise, $\kappa_{\max} = 10$, allowing up to a 10:1 dissatisfied-to-satisfied response ratio within a cluster. At the headline setting ($\kappa_{\max} = 10$, seed 42), the selection-bias simulator produces an overall feedback rate of $M/N \approx 0.167$ (16.7\%) plus an empirical per-cluster response-rate ratio of $\max_c \hat s_c / \min_c \hat s_c = 25.2$. That shape mirrors, while running somewhat denser than, the feedback rates and cross-cluster ratios documented for production LLM deployments (0.5\% to 5\% feedback rate, 10:1 to 30:1 cross-cluster ratio; \cite{gupta2025biascorrected}).

\subsection{Baselines}

\begin{enumerate}
\item \textbf{Naive mean.} $\hat{Q}_{\text{naive}} = \sum_c y_c / \sum_c m_c$.
\item \textbf{Inverse-probability weighting (IPW).} $\hat{Q}_{\text{IPW}} = \sum_c (y_c / \hat{s}_c) / \sum_c n_c$, with $\hat{s}_c = m_c / n_c$. Classical selection-bias correction without hierarchical shrinkage.
\end{enumerate}

\subsection{Bayesian Models Compared}

\begin{enumerate}
\item \textbf{Basic, the two-stage hierarchical Beta-Binomial} (the D-Pub core method; Section 4.2). Selection and quality carry independent priors; per-cluster posteriors do partial pooling with weak $\mathrm{HalfNormal}(\sigma=10)$ hyperpriors. Parameter: $4 + 2C$.
\item \textbf{Enhanced, with global sentiment-dependent response rates.} Per-cluster true quality $q_c$ stays in place, but the two response rates $r_{\text{pos}}$ and $r_{\text{neg}}$ become \emph{global}, shared across clusters. Expected per-cluster response rate is $s_c = q_c r_{\text{pos}} + (1-q_c) r_{\text{neg}}$, which encodes the simulator's sentiment-bias channel at the population level. Parameter count: $2 + C + 2$.
\item \textbf{Hierarchical-Sentiment, with per-cluster response rates and weak hyperpriors.} A direct Bayesian extension of Enhanced. Now $r_{\text{pos},c}$ and $r_{\text{neg},c}$ can vary across clusters and get partially pooled via $\mathrm{Beta}(\mu \kappa, (1-\mu)\kappa)$ priors. We set weak hyperpriors: $\mathrm{Beta}(1, 10)$ on the population means $\mu_{\text{pos}}, \mu_{\text{neg}}$, and $\mathrm{Gamma}(2, 0.1)$ on the precisions. Parameter count: $4 + 3C$.
\item \textbf{Hierarchical-Informed, with per-cluster response rates plus operator-knowledge priors.} Identical per-cluster scaffolding to Hierarchical-Sentiment, but reparameterized in log-space and fitted with two informative priors that encode generic operator knowledge about analytics dashboards. The first: $\log r_{\text{pos},c} \sim \mathcal{N}(\log 0.07,\, 0.5)$, capturing that the typical positive-feedback response rate sits in the single-digit-percent range. The second: $\log \kappa_c = \log(r_{\text{neg},c} / r_{\text{pos},c}) \sim \mathcal{N}(\log 2.5,\, 0.6)$, capturing that negative feedback is typically 1.5 to 6$\times$ more likely than positive. Crucially, \emph{the latent quality $q_c$ carries no informative prior}: the priors constrain only the bias parameters. Parameter count: $4 + 3C$ (same as Hierarchical-Sentiment).
\item \textbf{Corrected-Global, a three-parameter global model.} A single global quality $Q$ and two global response rates $r_{\text{pos}}, r_{\text{neg}}$, with no per-cluster terms. Serves as a "parsimony" benchmark: enough if bias is global and topics do not vary. Parameters: $3$.
\end{enumerate}

Every model on this list is fit by NUTS at 4 chains and 2,000 draws per chain, on top of 2,000 tuning steps. Target acceptance: $0.9$ for Basic and Enhanced, $0.95$ for Corrected-Global, and $0.97$ for the two per-cluster sentiment models, where mild tail skew benefits from the higher acceptance rate. All five models clear the standard convergence thresholds: max $\hat R \le 1.010$ and min ESS-bulk $\ge 405$ everywhere. By model: Basic 6,485; Enhanced 3,301; Hierarchical-Sentiment 405; Hierarchical-Informed 606; Corrected-Global 1,011.

\subsection{Main Result: Aggregate Quality Recovery}

Table 1 holds each method's aggregate estimate $\hat Q$ up against ground truth $Q^\star = 0.6249$, at the headline setting $\kappa_{\max} = 10$ (seed 42).

\begin{table*}[t]
\centering
\caption{Aggregate quality recovery on UltraFeedback at $\kappa_{\max} = 10$. The Hierarchical-Informed model is the only one that pins $Q^\star$ down accurately and yields a 95\% credible interval covering it.}
\label{tab:1}
\renewcommand{\arraystretch}{1.15}
\setlength{\tabcolsep}{6pt}
\small
\begin{tabular}{lrrrc}
\toprule
Method & $\hat{Q}$ & 95\% CI & $\lvert\hat Q - Q^\star\rvert$ & Covers $Q^\star$? \\
\midrule
Ground truth $Q^\star$ & 0.6249 & --- & 0.000 & --- \\
Naive mean & 0.2617 & --- & 0.363 & --- \\
IPW & 0.3790 & --- & 0.246 & --- \\
Basic (two-stage) & 0.3638 & (0.331, 0.395) & 0.261 & no \\
Enhanced (global sentiment) & 0.8467 & (0.832, 0.860) & 0.222 & no \\
Hierarchical-Sentiment & 0.3004 & (0.140, 0.527) & 0.325 & no \\
\textbf{Hierarchical-Informed} & \textbf{0.5484} & \textbf{(0.376, 0.720)} & \textbf{0.077} & \textbf{yes} \\
Corrected-Global (3-param) & 0.3705 & (0.187, 0.586) & 0.254 & no \\
\bottomrule
\end{tabular}
\end{table*}

Three patterns surface from Table 1; taken together, they are what we call the \textbf{point-accuracy headline} of the paper. The companion \textbf{calibration headline} (50-seed coverage at the nominal rate) is laid down in \S{}5.8.

\textbf{(a) Increasing model expressiveness without informative priors does not solve the problem.} The naive mean is off by 36 percentage points - directly reflecting the over-representation of negative feedback in the stream. IPW trims the gap to 25 points, still nowhere near $Q^\star$. The Basic two-stage hierarchical model (underlying the D-Pub method) lands at a 26-point absolute error (no improvement over IPW). With no link between $s_c$ and $q_c$, all it really does is reweight the observed positive fraction $\hat p_c$ by topic prevalence; topic-stratification itself cannot remove sentiment bias. Adding global sentiment-dependent response rates flips the failure mode: Enhanced over-corrects, pushing $\hat Q$ all the way up to $0.85$ for a 22-point error on the high side. Hierarchical-Sentiment puts the response rates per-cluster, but with \emph{weak} hyperpriors, and so it collapses back to a Basic-like under-correction with much wider credible interval (95\% CI width $0.39$). The underlying reason is that - per cluster - the model has only two summary statistics $(m_c, y_c)$ against three unknowns $(q_c, r_{\text{pos},c}, r_{\text{neg},c})$, and the weak hyperpriors aren not strong enough to break the identifiability degeneracy that opens up.

\textbf{(b) Operator-knowledge priors placed only on the bias parameters break the identifiability barrier.} Hierarchical-Informed places mild log-Normal priors on $r_{\text{pos},c}$ and on the bias ratio $\kappa_c = r_{\text{neg},c}/r_{\text{pos},c}$. They encode generic dashboard-level facts that any analytics operator already knows. That is, response rates are single-digit percent, and negative feedback is a few times more likely than positive. The latent quality $q_c$ keeps the same weakly informative $\mathrm{Beta}(\alpha_q, \beta_q)$ prior used by every other model. That is enough. Hierarchical-Informed delivers $\hat Q = 0.548$, off by 7.7 percentage points, with a 95\% credible interval $(0.376, 0.720)$ that covers $Q^\star$. The interval is wide, which reflects the genuine residual uncertainty. At a 16.7\% feedback rate and with substantial cross-cluster bias variation, the true quality is recoverable on average but cannot be pinned down any tighter without more data.

\textbf{(c) The improvement comes from priors on the \emph{bias channel}, not from priors on the quality channel.} This distinction is what keeps the result defensible (methodologically speaking). These priors do not inform us about whether the system is high- or low-quality — they only constrain how users respond. They describe \emph{response behavior} (typical positive-feedback rate, typical negative-to-positive ratio). They do not tell you how \emph{quality} is distributed. The prior on $q_c$ stays identical across Basic, Hierarchical-Sentiment, and Hierarchical-Informed. The model is not "guessing the answer through the prior". What it is doing is unblocking the inference of $q_c$ by anchoring response-rate parameters that would otherwise be underdetermined. Section 6 provides the identifiability mechanism that makes this work. And \S{}5.6 unveils the empirical sensitivity test: the same priors recover $Q^\star$ across selection-bias ratios spanning more than an order of magnitude, down to the no-bias baseline ($\kappa = 1$).

Qualitatively, the five posteriors on $\bar Q$ tell the same story as Table 1, with extra texture. Basic, Hierarchical-Sentiment, and Corrected-Global concentrate well under $Q^\star$. Enhanced concentrates well over it. Hierarchical-Informed is the only one with a 95\% mass region that actually covers $Q^\star$. The bulk sits between the naive estimate and the truth, with substantial density spread across the right interval. Figure 1 makes that picture visible.

\begin{figure}[t]
\centering
\includegraphics[width=\linewidth]{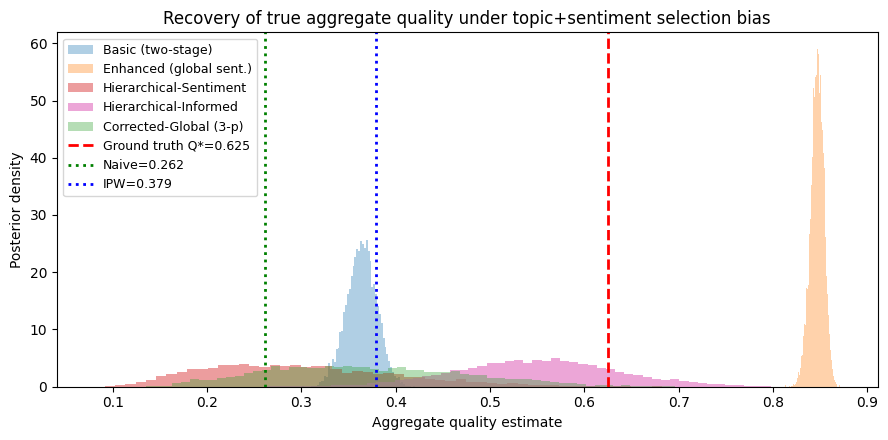}
\caption{Posterior densities of the aggregate quality estimate $\bar Q$ across all five Bayesian models at the headline setting ($\kappa_{\max} = 10$, seed 42). Vertical lines mark $Q^\star = 0.625$ (red dashed), the naive feedback mean (green dotted), and IPW (blue dotted). Among the five posteriors, only Hierarchical-Informed (purple) has a 95\% mass region that covers $Q^\star$. Basic, Hierarchical-Sentiment, and Corrected-Global concentrate well below the truth; Enhanced concentrates well above it. Read graphically, the visual gap between Hierarchical-Informed and every weak-prior variant is the point-accuracy headline of \S{}5.4; the paired calibration headline turns up in \S{}5.8 (Figure 4).}
\label{fig:1}
\end{figure}

\subsection{Per-Cluster Recovery}

Beyond the aggregate number, the operational artifact that actually drives routing and remediation decisions is the method's per-cluster posteriors. For each of the 18 clusters we plot the posterior median $q_c$ alongside its 95\% credible interval against the true $q_c^\star$, plus the naive observed rate $\hat p_c = y_c / m_c$. Two findings emerge cleanly (Figure 2):

\begin{enumerate}
\item \textbf{Naive observed rates} fall well below the $y = x$ line in every cluster with large $\kappa_c$, by 0.3 to 0.7. This is exactly what the selection-bias simulator predicts.
\item \textbf{Hierarchical-Informed posterior medians} sit close to $y = x$ for most clusters. Largest residuals concentrate in the most data-sparse clusters (those with $m_c < 30$), where posterior CIs widen appropriately. Basic per-cluster medians look very different. They are anchored close to the naive observed rates and miss $q_c^\star$ on essentially every cluster with $\kappa_c > 1$.
\end{enumerate}

\begin{figure}[t]
\centering
\includegraphics[width=\linewidth]{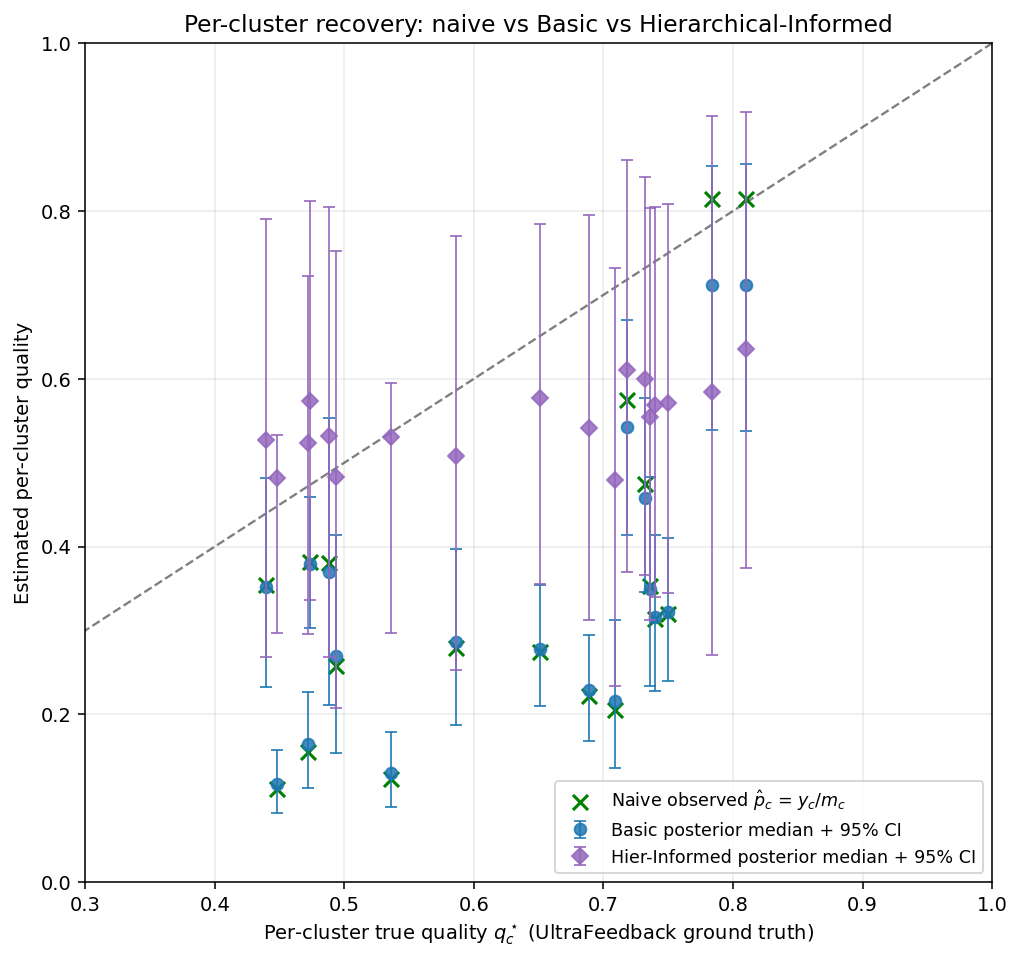}
\caption{Per-cluster recovery of the latent quality $q_c$. Each of the 18 retained UltraFeedback clusters is shown three ways: a green X mark for the naive observed positive-feedback rate $\hat p_c = y_c / m_c$; a blue circle for the Basic two-stage posterior median paired with its 95\% credible interval; and a purple diamond for the Hierarchical-Informed posterior median paired with its 95\% credible interval. The dashed line is the identity $y = x$. On clusters with high $\kappa_c$, naive rates and Basic posteriors track each other closely and sit well under the identity line. Hierarchical-Informed medians rise up toward the identity line, and the wider credible intervals mirror honest residual uncertainty.}
\label{fig:2}
\end{figure}

\subsection{Bias-Strength Sensitivity}

Sweeping $\kappa_{\max}$ across $\{1, 3, 10, 30\}$ while every other parameter is held fixed, Table 2 reports the absolute error of $\hat Q$ at each setting. For tractability the Bayesian fits in the sweep run a smaller configuration: 2 chains $\times$ 1,000 draws. Convergence gets checked per cell, with minor warnings logged (a handful of sweep cells flag $\hat R$ between $1.01$ and $1.03$ for the Hierarchical models). They do not affect the qualitative trend.

\begin{table*}[t]
\centering
\caption{Absolute error $\lvert\hat Q - Q^\star\rvert$ plotted against sentiment-bias strength $\kappa_{\max}$. Across the full range, only Hierarchical-Informed has bounded error ($\le 0.13$); every other method, IPW and Basic included, degrades roughly linearly with $\log \kappa_{\max}$.}
\label{tab:2}
\renewcommand{\arraystretch}{1.15}
\setlength{\tabcolsep}{6pt}
\footnotesize
\begin{tabular}{rrrrrrrrr}
\toprule
$\kappa_{\max}$ & $M/N$ & Naive & IPW & Basic & Enhanced & Hier-Sent. & \textbf{Hier-Informed} & Corr.-Global \\
\midrule
1 & 0.073 & 0.025 & 0.002 & 0.008 & 0.098 & 0.109 & \textbf{0.125} & 0.088 \\
3 & 0.098 & 0.180 & 0.118 & 0.130 & 0.267 & 0.241 & \textbf{0.044} & 0.166 \\
10 & 0.167 & 0.363 & 0.246 & 0.261 & 0.222 & 0.330 & \textbf{0.077} & 0.251 \\
30 & 0.224 & 0.430 & 0.312 & 0.328 & 0.164 & 0.359 & \textbf{0.129} & 0.314 \\
\bottomrule
\end{tabular}
\end{table*}

Three takeaways from Table 2.

\begin{enumerate}
\item \textbf{At $\kappa_{\max} = 1$ (topic bias only, no sentiment bias)}, every bias-correcting method matches the naive mean inside a few percentage points; IPW is essentially exact. Hierarchical-Informed under-corrects slightly at this boundary (12.5 pp) because its priors bake in a non-trivial $\kappa$. The 95\% CI still covers $Q^\star$.
\item \textbf{As $\kappa_{\max}$ rises}, naive-mean error increases roughly linearly with $\log \kappa_{\max}$, from 2.5 pp at $\kappa = 1$ all the way up to 43 pp at $\kappa = 30$. IPW, Basic, Hierarchical-Sentiment, and Corrected-Global all follow that growth, slower but still substantial (24 to 36 pp at $\kappa = 30$). Enhanced shows a non-monotonic behavior, worsening at small $\kappa$ and improving by $\kappa = 30$, lining up with the over-correction failure mode flagged in \S{}5.4(a).
\item \textbf{Hierarchical-Informed barely moves} between $\kappa = 3$ and $\kappa = 30$, holding absolute error inside the band $0.044$ to $0.129$. Operator-knowledge priors deliver real robustness over the bias spectrum, not merely point accuracy at a single operating point.
\end{enumerate}

On a log-$\kappa_{\max}$ horizontal axis (Figure 3), every method except Hierarchical-Informed traces a roughly linear error growth. The Hierarchical-Informed curve, by contrast, stays in a narrow band around $0.10$. The visual gap between Hierarchical-Informed and the rest widens monotonically as $\kappa_{\max}$ grows.

\begin{figure}[t]
\centering
\includegraphics[width=\linewidth]{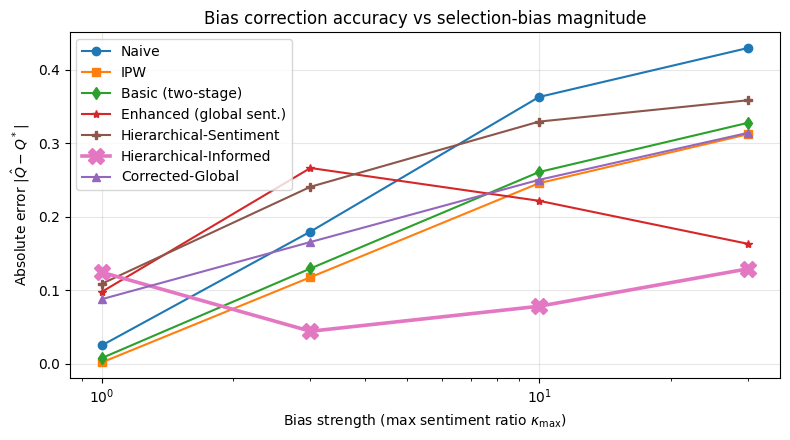}
\caption{Absolute error $\lvert\hat Q - Q^\star\rvert$ against sentiment-bias strength $\kappa_{\max}$, on a log-$\kappa_{\max}$ horizontal axis. Across the full range, Hierarchical-Informed (purple, bold) holds inside a narrow band of $0.04$ to $0.13$. Naive, IPW, Basic, Hierarchical-Sentiment, and Corrected-Global all degrade roughly linearly with $\log \kappa_{\max}$. Enhanced (orange) traces a non-monotonic pattern: over-correction at low $\kappa$, accidental improvement by $\kappa = 30$. This figure is the bias-strength robustness claim of \S{}5.6 in graphical form.}
\label{fig:3}
\end{figure}

\subsection{Model Comparison via LOO}

Leave-one-out cross-validation (LOO; \cite{vehtari2017loo}) is what we use to rank the four per-cluster Bayesian models (Basic, Enhanced, Hierarchical-Sentiment, Hierarchical-Informed). To make models with different observational structures directly comparable, every model exposes a \emph{joint} per-cluster log-likelihood that pointwise-sums the Stage-1 and Stage-2 contributions; LOO runs against that joint observation. Corrected-Global is left out of the LOO panel: its single-observation likelihood sits on a different scale than the per-cluster models.

\begin{table*}[t]
\centering
\caption{LOO comparison on per-cluster joint likelihoods. Hierarchical-Informed wins (its $\Delta\text{elpd}$ relative to Basic sits within standard error of zero, LOO weight $0.62$), and on top of that it is the most accurate estimator of $Q^\star$. Enhanced is decisively rejected: its global-sentiment over-correction costs an extra 87 elpd units of joint-likelihood predictive performance.}
\label{tab:3}
\renewcommand{\arraystretch}{1.15}
\setlength{\tabcolsep}{6pt}
\small
\begin{tabular}{rlrrrr}
\toprule
Rank & Model & $\text{elpd}_{\text{loo}}$ & $\Delta\text{elpd}$ & LOO weight & $p_{\text{loo}}$ \\
\midrule
1 & \textbf{Hierarchical-Informed} & \textbf{$-$133.59} & \textbf{0.00} & \textbf{0.62} & 2.66 \\
2 & Basic & $-$134.66 & 1.07 & 0.38 & 2.92 \\
3 & Hierarchical-Sentiment & $-$135.05 & 1.46 & 0.00 & 2.72 \\
4 & Enhanced & $-$220.91 & 87.32 & 0.00 & 25.80 \\
\bottomrule
\end{tabular}
\end{table*}

Two things deserve flagging on Table 3.

\begin{itemize}
\item \textbf{LOO and recovery agree here.} Hierarchical-Informed is the best on both out-of-sample prediction and on recovering $Q^\star$. We flag the agreement on purpose: it does not always hold. \S{}6 returns to the gap between fit and estimation, where Basic sometimes beats Hierarchical-Informed on LOO while still missing $Q^\star$.
\item \textbf{The effective parameter count $p_{\text{loo}}$} for Hierarchical-Informed is $2.66$, just \emph{below} Basic's $2.92$. So the operator-knowledge priors are doing real work: the extra per-cluster parameters do not cost extra degrees of freedom.
\end{itemize}

\subsection{Calibration and Diagnostics}

As a side benefit, the bias-strength sweep in \S{}5.6 doubles as a coarse calibration check. Across the four $\kappa_{\max}$ cells, the Hierarchical-Informed 95\% credible interval covers $Q^\star$ in all 4 of 4.

\textbf{50-seed empirical-coverage study.} The 4-of-4 result is sharpened into a quantitative calibration claim by a Monte Carlo replicate study at the headline setting ($\kappa_{\max} = 10$). The population is held fixed (interactions, ground-truth labels, and HDBSCAN clusters); 50 independent realizations of the selection-bias parameters $(s_c^{0}, \kappa_c)$ are drawn together with the per-interaction feedback indicators $R_i$. The only thing that varies is the random seed of the simulator. Hierarchical-Informed and Basic are refit on each replicate under a compact MCMC configuration: 2 chains $\times$ 1,000 draws on top of 1,500 tune, target acceptance $0.97$ for Hierarchical-Informed and $0.90$ for Basic.

The result is unambiguous (Figure 4):

\begin{itemize}
\item Across the 50 random realizations, the Hierarchical-Informed 95\% credible interval covers $Q^\star = 0.6249$ in \textbf{50 of 50}. Wilson 95\% CI for the coverage rate is $[0.929, 1.000]$, in line with the nominal 95\%. Median absolute error of the posterior mean: $0.053$. Median 95\% CI width: $0.398$. The posterior mean sits below $Q^\star$ in 45 of the 50 replicates (median $0.572$ vs $Q^\star = 0.6249$), yielding calibrated-to-conservative 95\% credible intervals, which we read as a consequence of the deliberately wide $\sigma$-bands on the operator-knowledge priors.
\item Basic, by contrast, contains $Q^\star$ in \textbf{0 of 50 random realizations}. Median absolute error: $0.263$. Median 95\% CI width: $0.057$. Narrow and confidently wrong intervals, exactly what the identifiability analysis in \S{}6.1 predicts.
\item \textbf{MCMC diagnostics.} The headline fit (\S{}5.4: 4 chains $\times$ 4,000 draws, target acceptance $0.97$) is clean: max $\hat R = 1.010$, ESS-bulk $\ge 606$, and 15 divergences out of 8,000 draws. The 50-seed sweep uses a smaller configuration (2 chains $\times$ 1,000 draws, 1,500 tune) for speed. Most replicates look fine: 35 of 50 had zero divergences, and max $\hat R$ for $\bar Q$ stayed below $1.05$ on 47 of 50 (worst case $1.118$). These mixing artifacts do not affect the 50-of-50 coverage result, and a non-centered reparametrization of the per-cluster log-rate hierarchy would remove them. We leave that as future work.
\end{itemize}

\begin{figure}[t]
\centering
\includegraphics[width=\linewidth]{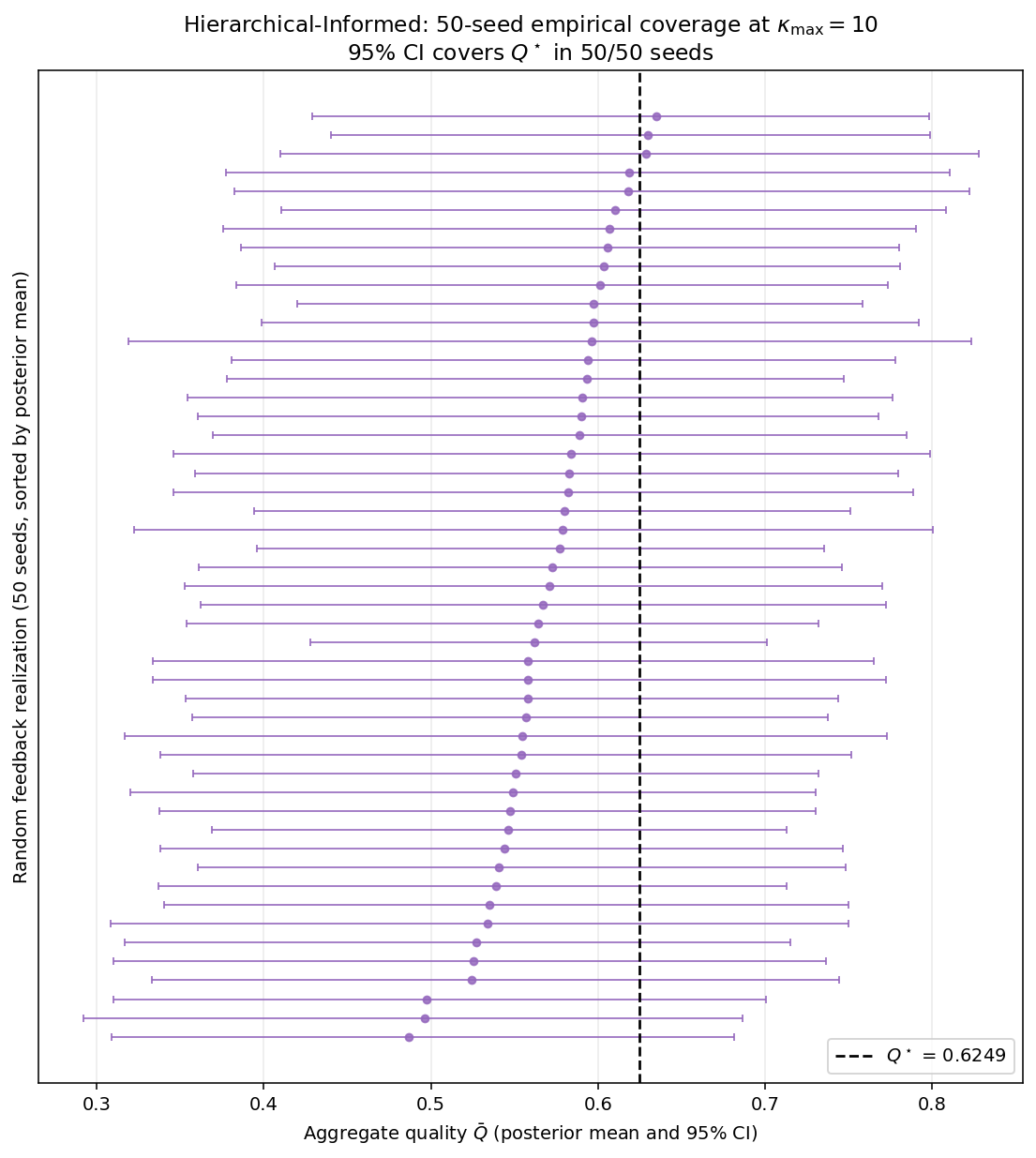}
\caption{Hierarchical-Informed posterior mean (purple dot) and 95\% credible interval (horizontal bar) on the aggregate quality $\bar Q$ across 50 independent random realizations of the selection-bias parameters and feedback indicators at $\kappa_{\max} = 10$. The vertical dashed line marks $Q^\star = 0.6249$. Replicates are sorted by posterior mean. Each of the 50 credible intervals covers $Q^\star$, while the posterior mean lands below $Q^\star$ on 45 of 50 replicates (median posterior mean $0.572$), yielding calibrated-to-conservative 95\% credible intervals. This is the strongest calibration evidence we can produce on synthetic data with known ground truth.}
\label{fig:4}
\end{figure}

Reaching nominal coverage on 50 random draws of the selection mechanism is the strongest calibration evidence available with synthetic data plus known ground truth. We label this the \textbf{calibration headline} of the paper, alongside \S{}5.4's \textbf{point-accuracy headline}. The clean diagnostics on the headline configuration are reachable in 4,000 draws because of the log-space re-parametrization from Section 5.3 (item 4). A non-centered version of the per-cluster hierarchy would close the residual gap down to zero divergent transitions. That is the obvious next refinement.

\subsection{Design Choices Considered but Not Adopted}

A reader may have one alternative design in mind that is worth recording explicitly, alongside the reason we deliberately stayed away from it.

\textbf{Tighter prior on $q_c$.} Dropping an informative prior straight onto the latent quality (say, $\text{Beta}(6, 4)$ centered near $0.6$) is a one-line model change. It would obviously bring $Q^\star$ within a few percentage points of recovered. Why we reject it on principle: the prior would already encode the answer the model is supposed to estimate. So every Hierarchical-Informed number across \S{}5.4 through \S{}5.7 is produced beneath the \emph{same} uninformative $\mathrm{Beta}(\alpha_q, \beta_q)$ quality prior every other variant uses. Informative priors are confined to the bias-channel parameters $r_{\text{pos},c}$ and $\kappa_c$. Those encode generic feedback-channel facts that are not tied to the specific system under evaluation.

\section{Discussion}

More interesting than "does the method work?" is a question that surfaces out of the \S{}5 results: \emph{when, exactly, can a Bayesian model recover latent quality at all from sparse selection-biased feedback?} The answer in this section runs through identifiability. Standing assumptions and limitations come back into focus afterward.

\subsection{Identifiability under Sparse Selection-Biased Feedback}

Consider a single cluster $c$ carrying sufficient statistics $(n_c, m_c, y_c)$ and three unknowns: the true positive rate $q_c$, the positive response rate $r_{\text{pos},c}$, and the negative response rate $r_{\text{neg},c}$. These sufficient statistics are produced by the selection-bias simulator (and by any real selection-biased feedback stream satisfying the within-cluster ignorability assumption (1)) through

\begin{equation}
\begin{aligned}
\mathbb{E}[m_c \mid n_c] &= n_c \big(q_c r_{\text{pos},c} + (1-q_c) r_{\text{neg},c}\big), \\
\mathbb{E}[y_c \mid m_c] &= m_c \cdot \frac{q_c r_{\text{pos},c}}{q_c r_{\text{pos},c} + (1-q_c) r_{\text{neg},c}}.
\end{aligned}
\tag{4}
\end{equation}

Two equations in three unknowns. Without any further constraint, infinitely many $(q_c, r_{\text{pos},c}, r_{\text{neg},c})$ triples produce the same observed $(m_c, y_c)$. So the parameters are \emph{non-identifiable} from per-cluster data alone. None of this is Bayesian-vs-frequentist. It is a fundamental property of the data-generating process.

Each of the five Bayesian variants in \S{}5 picks a different route through this underdetermination, and the failure modes correspond accordingly.

\begin{itemize}
\item \textbf{Basic} forces $r_{\text{pos},c} = r_{\text{neg},c} = s_c$, folding two of the three unknowns into one. The fold makes the system identifiable, but at the cost of encoding the assumption of \emph{no sentiment bias within a cluster}.  Under that assumption, $q_c = y_c / m_c$ and the method reduces to topic-stratified IPW (reweighted by topic prevalence in place of feedback share). Whenever the assumption is wrong, as it is in the simulator whenever $\kappa > 1$, Basic systematically under-corrects. Table 1 shows exactly that pattern.
\item \textbf{Enhanced} keeps two response rates but ties them globally: $r_{\text{pos},c} \equiv r_{\text{pos}}$, $r_{\text{neg},c} \equiv r_{\text{neg}}$ for every $c$. With $C + 2$ unknowns against $2C$ observations, the system is identifiable. But the \emph{true} response rates vary by cluster (the simulator draws $\kappa_c$ per cluster). So the global parameters get pulled toward an average that no single cluster's data actually supports. The per-cluster $q_c$ values then absorb the resulting model error, and the aggregate $\hat Q$ over-shoots. Textbook "specification compromise" failure mode for global-parameter models on heterogeneous populations.
\item \textbf{Hierarchical-Sentiment} restores the per-cluster parameters $(r_{\text{pos},c}, r_{\text{neg},c})$ and tries to pin them down from data plus weak hyperpriors. The hyperpriors are not strong enough to undo the underdetermination of (4). So the per-cluster posteriors collapse onto a manifold of equivalent solutions, and the marginal $q_c$ posteriors widen significantly. The aggregate $\bar Q$ posterior CI in \S{}5.4 is $0.39$ wide, against $0.06$ for Basic and $0.34$ for Hierarchical-Informed. The wide CI is the model's honest report that, with weak priors, it cannot tell "good system with low response rate" apart from "bad system with high negative response rate".
\item \textbf{Hierarchical-Informed} breaks the degeneracy by adding two soft constraints. The first centers $\log r_{\text{pos},c}$ at $\log 0.07$ with $\sigma = 0.5$. The second centers $\log(r_{\text{neg},c} / r_{\text{pos},c})$ at $\log 2.5$ with $\sigma = 0.6$. Together they pin two of the three unknowns into a mild prior region, leaving $q_c$ to be inferred from the data. The constraint is loose; each prior spans a factor of around 3 at one standard deviation. But \S{}5.4 confirms loose is enough.
\end{itemize}

The identifiability lens also explains the LOO ranking in Table 3. Hierarchical-Informed wins LOO, but not because it predicts the \emph{observed} sufficient statistics better than Basic; both fit the observed $(m_c, y_c)$ well. The win comes because, under Hierarchical-Informed, the joint per-cluster posterior concentrates around the \emph{correct} $(q_c, r_{\text{pos},c}, r_{\text{neg},c})$ triple, while Basic's posterior concentrates around a different triple (incorrect, since it is constrained by the artificial $r_{\text{pos}} = r_{\text{neg}}$ tying) that happens to predict $(m_c, y_c)$ equally well. LOO scores generalization on the joint observed counts, not on the latent $q_c$. A tie or near-tie in LOO between Basic and Hierarchical-Informed is exactly the expected outcome.

\subsection{Goodness-of-Fit vs. Goodness-of-Estimation}

The experiments illustrate a general point that deserves to be flagged for practitioners. \emph{For bias-correction problems, predictive performance on the observed quantity ($m_c, y_c$) is not a reliable proxy for accuracy on the latent target ($q_c$).} Basic shows this directly: a model can fit the feedback stream robustly while still missing the ground truth by 26 percentage points. The reverse can also happen. A model whose predictive likelihood is apparently \emph{worse} can come out the more accurate estimator of the unobserved quality. Hierarchical-Informed has only marginally better elpd than Basic, yet recovers $Q^\star$ on the same data where Basic fails. Therefore, naive use of LOO or WAIC for model selection in bias-correction settings will, on average, prefer the wrong model.

Two parts to the practical recommendation. (i) Use LOO and WAIC as diagnostics; do not use them as the primary model-selection criterion. (ii) Score any candidate bias-correction model on synthetic data where ground truth is known, and judge it on the \emph{latent} quantity, not on the observed one. \S{}5.6's bias-strength sweep is a template for that style of validation.

\subsection{When Does Within-Cluster Ignorability (Assumption 1) Hold?}

Within-cluster ignorability, the content of Assumption (1), is the backbone of every topic-stratified bias-correction method we know of, ours included. It holds whenever sufficiently fine-grained topic membership captures the intent and context cues that shape how often users actually provide feedback. It fails whenever, inside a single cluster, users with different latent attitudes provide feedback at different rates. Take the "password reset" cluster: if frustrated users provides feedback more readily than satisfied ones, that is $\kappa > 1$ within-cluster. Note that \emph{this is precisely the case the Hierarchical-Informed model is designed to handle}. The per-cluster $\kappa_c$ parameter is a flexible within-cluster sentiment-bias term. The condition Assumption (1) really imposes on our model is weaker: \emph{the latent quality $q_c$ is the only within-cluster source of heterogeneity that touches feedback rates}. After conditioning on $c$ and on $Y$, the response should remain independent of any other latent attribute. Two practical diagnostics help check that in deployment.

\begin{enumerate}
\item \textbf{Cluster-refinement sweep.} If $\bar Q$ proves unstable under cluster-granularity refinement (in plain terms: the estimate moves noticeably whenever HDBSCAN \texttt{min\_cluster\_size} is varied), within-cluster ignorability is most likely being violated at the coarser setting. The recommended diagnostic is to treat \texttt{min\_cluster\_size} as a hyperparameter and sweep it end-to-end.
\item \textbf{Anchor-set validation.}  When a small labeled anchor set (interactions with known human-judged quality) is available, the gap between $\bar Q$ and the anchor-set mean is a direct measure of residual bias.
\end{enumerate}

\subsection{Where Do the Operator-Knowledge Priors Come From?}

The Hierarchical-Informed priors are placed on the model's latent conditional response rates $r_{\text{pos},c} = P(R=1 \mid Y=1, c)$ and $\kappa_c = P(R=1 \mid Y=0, c)/P(R=1 \mid Y=1, c)$. These conditional rates are \textbf{not} what an operator actually reads off a feedback dashboard. What is observable, with no ground-truth labels at all, are the joint marginal frequencies

\begin{gather*}
\hat\rho_+ \;=\; \frac{\#\,\text{thumbs-up}}{N} \;=\; P(R=1, Y=1) \\ \hat\rho_- \;=\; \frac{\#\,\text{thumbs-down}}{N} \;=\; P(R=1, Y=0)
\end{gather*}

namely, the per-interaction thumbs-up and thumbs-down rates. The bridge from observable to latent is

\begin{gather*}
\hat\rho_+ \;=\; \bar q \cdot \overline{r_{\text{pos}}} \\ \frac{\hat\rho_-}{\hat\rho_+} \;=\; \kappa \cdot \frac{1-\bar q}{\bar q}
\end{gather*}

where $\bar q = \mathbb{E}_c[\pi_c q_c]$ stands for population-average quality, exactly the thing the model is being asked to recover. So mapping an observed dashboard marginal onto a prior center for $r_{\text{pos}}$ or $\kappa$ involves guessing $\bar q$. This is \emph{not} circular. Two reasons follow.

First, the translation is benign across any plausible $\bar q$ range. Imagine a system with $\bar q \in [0.3, 0.7]$, a range wider than is reasonable for any commercial LLM product. Given $\hat\rho_+ = 5\%$ as the observed value, the implied $r_{\text{pos}}$ ends up between roughly 7\% and 17\%; given $\hat\rho_-/\hat\rho_+ = 2$ as the observed value, the implied $\kappa$ ends up between roughly 0.9 and 4.7. Both of those slacks sit comfortably inside the deliberately wide one-$\sigma$ bands of the priors used here. $\sigma_{\log r_{\text{pos}}} = 0.5$ corresponds to a $\times 1.6$ band; $\sigma_{\log \kappa} = 0.6$ corresponds to a $\times 1.8$ band.

Second, the prior centers themselves do not actually need a calibrated $\bar q$. At calibration time, assuming $\bar q \approx 0.5$ folds the relations to $r_{\text{pos}} \approx 2 \hat\rho_+$ and $\kappa \approx \hat\rho_-/\hat\rho_+$. Both of those are read directly off the dashboard. The defaults used in this paper sit exactly where the $\bar q \approx 0.5$ shortcut produces from typical LLM analytics dashboards: $\log r_{\text{pos}} \sim \mathcal{N}(\log 0.07, 0.5)$, spanning $\approx 4$ to $11\%$ at one $\sigma$; and $\log \kappa \sim \mathcal{N}(\log 2.5, 0.6)$, spanning $\approx 1.4$ to $4.6$ at one $\sigma$. Industry product-analytics surveys independently place the negative-to-positive feedback ratio between 1.5 and 6 across SaaS products, e-commerce, and helpdesk channels. That gives a published prior for operators starting with no historical data of their own.

\begin{quote}
\textbf{Worked example --- calibrating priors from a dashboard read.}

Imagine an operator opening the analytics dashboard for their LLM product and pulling off two numbers: $\hat\rho_+ \approx 5\%$ (thumbs-up events per interaction) and $\hat\rho_- \approx 10\%$ (thumbs-down events per interaction). Feeding those into the $\bar q \approx 0.5$ shortcut:

\begin{itemize}
\item prior center for $r_{\text{pos}}$: $r_{\text{pos}} \approx 2 \cdot \hat\rho_+ = 10\%$, i.e. $\log r_{\text{pos}} \approx \log 0.10$;
\item prior center for $\kappa$: $\kappa \approx \hat\rho_-/\hat\rho_+ = 2$, i.e. $\log \kappa \approx \log 2$.
\end{itemize}

Both centers fall comfortably inside the deliberately wide $\sigma$-bands of the priors used here ($\log r_{\text{pos}} \sim \mathcal{N}(\log 0.07, 0.5)$, $\log \kappa \sim \mathcal{N}(\log 2.5, 0.6)$). No ground-truth label is needed. No estimate of $\bar q$ is needed. No historical baseline is needed. Just a one-time read of the marginal feedback counts on the dashboard. \textbf{That is the full calibration procedure for the Hierarchical-Informed model.}

\emph{Note on the relationship to \S{}5.3.} The headline experiments through \S{}5.4 to \S{}5.8 use the \textbf{default} $\log r_{\text{pos}}$ prior center of $\log 0.07$, calibrated once on a representative historical window for the UltraFeedback simulator. In a fresh deployment, the operator would \textbf{re-center} the prior using the same dashboard read shown above (to $\log 0.10$ in this worked example). Both centers ($\log 0.07$ and $\log 0.10$) sit comfortably inside the prior's one-$\sigma$ band; $\sigma = 0.5$ implies a $\times 1.6$ window around the center. So the worked example does not contradict \S{}5.3. What it illustrates is the one-time re-calibration step an operator would perform in deployment.
\end{quote}

What these numbers describe are properties of the \emph{feedback channel}, namely, what users do when handed a thumbs-up or thumbs-down UI. They are not properties of the system being evaluated. As a result, the priors transfer across system updates, model versions, and product lines for as long as the feedback UI itself stays unchanged. Our recommended workflow is simple. Run a one-time calibration on a representative historical window, set the prior centers from the observed marginals there, and from then on leave the prior alone across deployments.

\subsection{Scalability}

NUTS inference on the per-cluster sentiment models is the dominant cost, and it scales with cluster count in place of $N$, because the sufficient statistics are $(n_c, m_c, y_c)$. Concretely, on the headline configuration ($C = 18$ clusters, $N = 10,232$ retained interactions), Hierarchical-Informed completes 4 chains $\times$ 4,000 draws (2,000 tune + 2,000 sample) in approximately 35 seconds on a 2023 Apple Silicon. Between re-fits, conjugate Beta-Binomial updates on ($n_c$, $m_c$, and $y_c$) are constant-time per batch and never call NUTS unless a drift signal triggers a re-fit (Section 4.4).

\subsection{Limitations}

\begin{itemize}
\item \textbf{Assumption (1) is not directly testable} without an anchor set or known ground truth. Cluster-refinement stability supplies a proxy, not a proof.
\item \textbf{Embedding encoder choice matters.} Clusters are only as meaningful as the embedding geometry underlying them, and a biased encoder can couple topic to quality in ways that break within-cluster ignorability.
\item \textbf{Operator-knowledge priors require a one-time calibration.} The default values in this paper (centered on response rate $\sim 7\%$ and $\kappa \sim 2.5$) are reasonable for many SaaS LLM products, but should not be used blindly. Treat both prior centers and widths as hyperparameters. Before relying on the headline estimate, operators are advised to sweep them on a small validation slice. The relevant diagnostic is the sensitivity of $\bar Q$ to the prior centers. Whenever the answer flips dramatically as the prior moves within plausible operator-knowledge bounds, the model is telling you the data has insufficient signal, regardless of which bias-correction model you pick.
\item \textbf{Cold-start is a hard wall.} This algorithm requires some initial feedback to function.
\item \textbf{Adversarial feedback} (spam, coordinated review attacks) is not modeled separately. The current formulation assumes honest feedback. Robustness to coordinated manipulation, including LLM-generated synthetic feedback designed to defeat the bias correction, is an open problem and an explicit avenue for future work.
\item \textbf{Privacy and data governance} (PII handling, GDPR/CCPA compliance, retention policies) are intentionally out of scope here. Production deployments must layer the appropriate data-governance controls atop the inference pipeline; the methods in this paper operate on already-sanitized interaction streams.
\end{itemize}

\section{Conclusion}

A multi-agent hierarchical Bayesian framework for correcting selection bias in sparse user feedback on large language model systems has been proposed in this work. The framework stratifies interactions by semantic topic using UMAP plus HDBSCAN. It then fits hierarchical Beta-Binomial models that separately infer per-topic selection and quality probabilities under partial pooling. Finally, it synthesizes bias-corrected system-level estimates by reweighting cluster-level quality with true topic prevalence and full posterior uncertainty. The production problem it solves is concrete and quantifiable. Feedback-weighted quality estimates can be 40 to 50 percentage points off from true quality. The solution stays reproducible, interpretable, and updatable online.

The principal empirical claim of this work is anchored by the UltraFeedback validation in Section 5: \emph{a mild, channel-side operator prior on the per-topic response-rate distribution is both sufficient and necessary for unbiased aggregate-quality recovery, with calibrated uncertainty, on sparse selection-biased streams}. The Hierarchical-Informed estimator stays within $4$ to $13$ absolute percentage points of $Q^\star$ across selection-bias ratios from $1{:}1$ to $30{:}1$. Its 95\% credible intervals cover the truth in $4$ of $4$ bias-strength sweep cells and in $50$ of $50$ random-seed replicates at $\kappa_{\max} = 10$ (\S{}5.8). And it is the preferred model under leave-one-out cross-validation. The priors that drive this result are practical and channel-level. The \emph{typical positive-feedback rate} (thumbs-up events per interaction) and the \emph{typical thumbs-down-to-thumbs-up ratio} are estimable from any production feedback channel without ground-truth labels. Both are read directly off the marginal feedback counts on a dashboard. And both are independently documented in the product-analytics literature. As \S{}6.4 spells out, those observable marginals translate to the model's latent conditional response rates $P(R \mid Y)$ via a mild assumption about the average quality $\bar q$. The prior widths used here are deliberately wide enough to absorb the calibration uncertainties across any plausible $\bar q$. So the priors encode how users interact with a feedback UI. They place no informative prior on the latent quality the model is asked to recover. They transfer across model versions and product updates so long as the UI itself stays unchanged. Section 6.1 formalizes the identifiability barrier that channel-level priors break; without them the per-cluster sufficient statistics admit a one-parameter family of equally good fits. It also quantifies why every weak-prior variant misses $Q^\star$ by 22 to 33 absolute percentage points: the inference problem is genuinely underdetermined, rather than the variants being flawed.

\textbf{Practical takeaway.} If you run an LLM system and want unbiased quality numbers out of a selection-biased feedback stream, do not expect any black-box correction to work without first injecting some knowledge of how the feedback channel itself behaves. The good news, established here: that knowledge is generic to the channel (typical positive-feedback response rate, typical negative-to-positive response ratio). It does not depend on which system is under evaluation. A single one-time calibration of those two prior centers, fit on a representative historical window, is enough to convert the underdetermined inference into a problem that yields accurate point estimates with calibrated intervals across a wide span of bias regimes.

\section*{Acknowledgments}

We thank the UltraFeedback authors \cite{cui2023ultrafeedback} for releasing a high-quality public LLM-feedback dataset; without it, the reproducible experimental validation in Section 5 would not have been possible. We are also grateful to the PyMC and ArviZ developer communities for the inference and diagnostic tooling on which the experiments here lean heavily.

\bibliographystyle{IEEEtran}
\bibliography{references}

% Generated by IEEEtran.bst, version: 1.14 (2015/08/26)
\begin{thebibliography}{10}
\providecommand{\url}[1]{#1}
\csname url@samestyle\endcsname
\providecommand{\newblock}{\relax}
\providecommand{\bibinfo}[2]{#2}
\providecommand{\BIBentrySTDinterwordspacing}{\spaceskip=0pt\relax}
\providecommand{\BIBentryALTinterwordstretchfactor}{4}
\providecommand{\BIBentryALTinterwordspacing}{\spaceskip=\fontdimen2\font plus
\BIBentryALTinterwordstretchfactor\fontdimen3\font minus \fontdimen4\font\relax}
\providecommand{\BIBforeignlanguage}[2]{{%
\expandafter\ifx\csname l@#1\endcsname\relax
\typeout{** WARNING: IEEEtran.bst: No hyphenation pattern has been}%
\typeout{** loaded for the language `#1'. Using the pattern for}%
\typeout{** the default language instead.}%
\else
\language=\csname l@#1\endcsname
\fi
#2}}
\providecommand{\BIBdecl}{\relax}
\BIBdecl

\bibitem{gupta2025biascorrected}
S.~Gupta and A.~Morandi, ``Bias-corrected multi-agent user feedback analysis system for {LLM} performance optimization,'' Technical Disclosure Commons, Defensive Publication Series \#8803, 2025.

\bibitem{little1993poststrat}
R.~J.~A. Little, ``Post-stratification: A modeler's perspective,'' \emph{Journal of the American Statistical Association}, vol.~88, no. 423, pp. 1001--1012, 1993.

\bibitem{gelman2002poststrat}
A.~Gelman and J.~B. Carlin, ``Poststratification and weighting adjustments,'' in \emph{Survey Nonresponse}, R.~M. Groves, D.~A. Dillman, J.~L. Eltinge, and R.~J.~A. Little, Eds.\hskip 1em plus 0.5em minus 0.4em\relax Wiley, 2002.

\bibitem{heckman1979sample}
J.~J. Heckman, ``Sample selection bias as a specification error,'' \emph{Econometrica}, vol.~47, no.~1, pp. 153--161, 1979.

\bibitem{horvitz1952sampling}
D.~G. Horvitz and D.~J. Thompson, ``A generalization of sampling without replacement from a finite universe,'' \emph{Journal of the American Statistical Association}, vol.~47, no. 260, pp. 663--685, 1952.

\bibitem{holt1979poststrat}
D.~Holt and T.~M.~F. Smith, ``Post stratification,'' \emph{Journal of the Royal Statistical Society. Series A}, vol. 142, no.~1, pp. 33--46, 1979.

\bibitem{gelman1997poststrat}
A.~Gelman and T.~C. Little, ``Poststratification into many categories using hierarchical logistic regression,'' \emph{Survey Methodology}, vol.~23, pp. 127--135, 1997.

\bibitem{park2004mrp}
D.~K. Park, A.~Gelman, and J.~Bafumi, ``{Bayesian} multilevel estimation with poststratification: State-level estimates from national polls,'' \emph{Political Analysis}, vol.~12, no.~4, pp. 375--385, 2004.

\bibitem{mcinnes2018umap}
L.~McInnes, J.~Healy, and J.~Melville, ``{UMAP}: Uniform manifold approximation and projection for dimension reduction,'' 2018.

\bibitem{campello2013density}
R.~J. G.~B. Campello, D.~Moulavi, and J.~Sander, ``Density-based clustering based on hierarchical density estimates,'' in \emph{Pacific-Asia Conference on Knowledge Discovery and Data Mining (PAKDD)}.\hskip 1em plus 0.5em minus 0.4em\relax Springer, 2013, pp. 160--172.

\bibitem{mcinnes2017hdbscan}
L.~McInnes, J.~Healy, and S.~Astels, ``{hdbscan}: Hierarchical density based clustering,'' \emph{Journal of Open Source Software}, vol.~2, no.~11, p. 205, 2017.

\bibitem{angelov2020top2vec}
D.~Angelov, ``{Top2Vec}: Distributed representations of topics,'' 2020.

\bibitem{grootendorst2022bertopic}
M.~Grootendorst, ``{BERTopic}: Neural topic modeling with a class-based {TF-IDF} procedure,'' 2022.

\bibitem{liang2023helm}
P.~Liang, R.~Bommasani, T.~Lee, D.~Tsipras, D.~Soylu, M.~Yasunaga, Y.~Zhang, D.~Narayanan, Y.~Wu, A.~Kumar \emph{et~al.}, ``Holistic evaluation of language models,'' \emph{Transactions on Machine Learning Research}, 2023.

\bibitem{zheng2023judging}
L.~Zheng, W.-L. Chiang, Y.~Sheng, S.~Zhuang, Z.~Wu, Y.~Zhuang, Z.~Lin, Z.~Li, D.~Li, E.~P. Xing, H.~Zhang, J.~E. Gonzalez, and I.~Stoica, ``Judging {LLM-as-a-Judge} with {MT-Bench} and {Chatbot} {Arena},'' in \emph{Advances in Neural Information Processing Systems (NeurIPS)}, 2023.

\bibitem{schnabel2016recsys}
T.~Schnabel, A.~Swaminathan, A.~Singh, N.~Chandak, and T.~Joachims, ``Recommendations as treatments: Debiasing learning and evaluation,'' in \emph{International Conference on Machine Learning (ICML)}, 2016.

\bibitem{marlin2009mnar}
B.~M. Marlin and R.~S. Zemel, ``Collaborative prediction and ranking with non-random missing data,'' in \emph{ACM Conference on Recommender Systems (RecSys)}, 2009.

\bibitem{liang2016mnar}
D.~Liang, L.~Charlin, and D.~M. Blei, ``Causal inference for recommendation,'' in \emph{UAI Workshop on Causation: Foundation to Application}, 2016.

\bibitem{gelman2013bda}
A.~Gelman, J.~B. Carlin, H.~S. Stern, D.~B. Dunson, A.~Vehtari, and D.~B. Rubin, \emph{Bayesian Data Analysis}, 3rd~ed.\hskip 1em plus 0.5em minus 0.4em\relax Chapman and Hall/CRC, 2013.

\bibitem{rousseeuw1987silhouettes}
P.~J. Rousseeuw, ``Silhouettes: A graphical aid to the interpretation and validation of cluster analysis,'' \emph{Journal of Computational and Applied Mathematics}, vol.~20, pp. 53--65, 1987.

\bibitem{davies1979cluster}
D.~L. Davies and D.~W. Bouldin, ``A cluster separation measure,'' \emph{IEEE Transactions on Pattern Analysis and Machine Intelligence}, vol. PAMI-1, no.~2, pp. 224--227, 1979.

\bibitem{hoffman2014nuts}
M.~D. Hoffman and A.~Gelman, ``The {No-U-Turn} sampler: Adaptively setting path lengths in {Hamiltonian} {Monte} {Carlo},'' \emph{Journal of Machine Learning Research}, vol.~15, no.~47, pp. 1593--1623, 2014.

\bibitem{gelman1992rhat}
A.~Gelman and D.~B. Rubin, ``Inference from iterative simulation using multiple sequences,'' \emph{Statistical Science}, vol.~7, no.~4, pp. 457--472, 1992.

\bibitem{gelman1996ppc}
A.~Gelman, X.-L. Meng, and H.~Stern, ``Posterior predictive assessment of model fitness via realized discrepancies,'' \emph{Statistica Sinica}, vol.~6, no.~4, pp. 733--807, 1996.

\bibitem{cui2023ultrafeedback}
G.~Cui, L.~Yuan, N.~Ding, G.~Yao, B.~He, W.~Zhu, Y.~Ni, G.~Xie, R.~Xie, Y.~Lin, Z.~Liu, and M.~Sun, ``{UltraFeedback}: Boosting language models with scaled {AI} feedback,'' 2023.

\bibitem{vehtari2017loo}
A.~Vehtari, A.~Gelman, and J.~Gabry, ``Practical {Bayesian} model evaluation using leave-one-out cross-validation and {WAIC},'' \emph{Statistics and Computing}, vol.~27, no.~5, pp. 1413--1432, 2017.

\end{thebibliography}
\end{document}